\documentclass[10pt,a4paper]{article}
\usepackage[margin=23mm]{geometry}
\usepackage[T1]{fontenc}
\usepackage{lmodern,amsmath,amssymb,booktabs,tabularx,graphicx,placeins,xcolor}
\usepackage[numbers,sort&compress]{natbib}
\usepackage[colorlinks=true,linkcolor=teal,citecolor=teal,urlcolor=teal]{hyperref}
\graphicspath{{figures/}}
\newcolumntype{Y}{>{\raggedright\arraybackslash}X}
\title{GREMLIN: Generating Random Environments\\for Meta-Learning with Interacting Networks}
\author{Octave Gaspard\,\raisebox{3pt}{\href{https://orcid.org/0009-0000-7375-3538}{%
    \includegraphics[height=7pt]{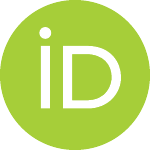}}}\\[2pt]
  \normalsize Independent Researcher}
\date{November 2026}
\begin{document}
\maketitle
\begin{abstract}
Can randomly parameterized neural environments supply useful experience for
learning a reinforcement learning algorithm? We investigate this question using
DiscoRL's rule-discovery framework, replacing human-designed training games
with sixteen selected neural environments. Each game combines bounded vector
dynamics with a random strategy-based reward and an independent outcome reward.
After 224.8 million agent--environment steps, we froze the teacher and trained
fresh agents. DQN outscored GREMLIN on most games, both generated and
human-designed. GREMLIN nevertheless outperformed the untrained teacher on all
26 games, showing that generated experience can improve a learning rule.
It remains unclear whether this approach can yield a competitive rule with
longer meta-training.
\end{abstract}

\section{Introduction}
DiscoRL \cite{discorl} learns reinforcement learning algorithms by training agents
on a collection of games. GREMLIN asks whether random neural networks could
supply those games: a potentially unlimited source of environments for learning
how to learn. The experiment reported here is narrower: one learning rule trained
on sixteen selected random neural environments.

Ordinary algorithms prescribe how experience changes an agent's network.
DQN \cite{dqn} trains estimates of future reward towards a
reward-plus-next-state-value target. DiscoRL instead learns a \emph{teacher}:
a neural rule that reads actions, rewards and agent predictions, then produces
desired action probabilities and prediction outputs. Agent updates move outputs
towards these targets. Differentiating through those updates trains the teacher
to increase agent reward. This outer training is \emph{discovery}, or meta-training.
To evaluate a teacher, we freeze it and train new agents from scratch.

The released Disco103 rule \cite{code} was discovered across 103 human-designed
games. Its weights stay fixed while fresh agents learn, potentially with
different observations, action counts and architectures. We use that
implementation and Disco103 as a reference; GREMLIN starts from random teacher
weights. We use low-dimensional states and small dense agents because our
budget did not allow image-based meta-training with convolutional agents like
those used in DiscoRL.

We ask whether training a teacher on neural games improves its ability to
teach agents on human-designed tasks. Testing fresh agents on the sixteen
training games measures learning within the bank; testing them on six separate
human games measures transfer. With one discovery run, we can demonstrate a
proof of concept but cannot establish how reliably discovery succeeds or
whether it can produce a competitive rule.

\paragraph{Related work.}
Learned Policy Gradient (LPG) \cite{lpg} is an earlier method for learning an RL
update rule from simple, human-specified task families, including grid worlds,
and transferring it to Atari. This is a precedent for learning a transferable
rule, not for our environment generator. GROOVE \cite{groove} directly studies
environment design for rule discovery: it adapts grid-world curricula using
algorithmic regret, a performance gap relative to a reference algorithm.
Our environments instead have randomly sampled neural dynamics and rewards,
and are screened once before discovery rather than adapted to the evolving rule.
Procgen \cite{procgen} randomizes
levels within human-designed games; GREMLIN randomizes transition and reward
functions within a specified neural family. Learned synthetic environments
\cite{synthetic} are another distinct approach: their networks are optimized
for target tasks, whereas ours are sampled and screened, not fitted.

\begin{figure}[ht]
\centering
\includegraphics[width=\linewidth]{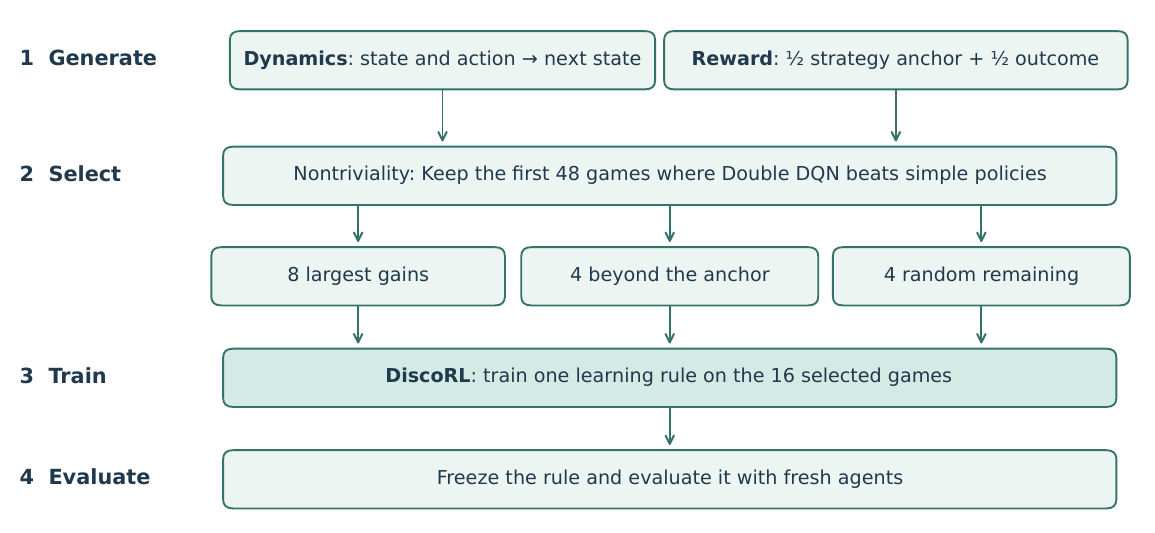}
\caption{GREMLIN workflow.}
\end{figure}

\section{Random games with deliberate structure}
Each game has a state $s\in[-1,1]^{32}$, four discrete actions, and exactly
128 steps per episode. The observation is $(s_t,1-t/128)$ for
$t=0,\ldots,127$, giving 33 inputs.

\subsection{Dynamics and sampling}
Action $a$ selects an eight-dimensional embedding $e_a$. State, action embedding
and fresh noise are projected separately into the first hidden layer:
\begin{align*}
\epsilon_t&\sim\mathcal N(0,I_{32}),\\
h_t^{(1)}&=\tanh(g_sW_ss_t+g_aW_ae_{a_t}+\sigma W_\epsilon\epsilon_t),\\
h_t^{(2)}&=\tanh(g_hW_hh_t^{(1)}),\\
u_t&=\tanh(W_oh_t^{(2)}),\\
s_{t+1}&=(1-\alpha)s_t+\alpha u_t.
\end{align*}
Both hidden layers have width 128; the proposal has 32 coordinates.
The last tanh and linear interpolation keep the state in $[-1,1]^{32}$.
For every dynamics, strategy and outcome network, weights with $n$ inputs have
independent entries $W_{ij}\sim\mathcal N(0,1/n)$ and zero biases. In the first
dynamics layer, $n=32$ for state and noise and $n=8$ for the action embedding.
Each action
embedding has eight independent standard-normal coordinates.

Gains are drawn once per game (see Table~\ref{tab:dynamics}). Log-uniform on $[l,u]$ means sampling $\ln x$
uniformly, equivalently density $p(x)=1/[x\ln(u/l)]$ on that interval.

\begin{table}[ht]\centering\small
\begin{tabular}{ll}\toprule
Quantity & Distribution\\\midrule
Inertia coefficient $\alpha$ & Log-uniform on $[1/16,1]$\\
State and hidden gains $g_s,g_h$ & Each log-uniform on $[0.5,2]$\\
Action gain $g_a$ & Log-uniform on $[0.25,4]$\\
Noise gain $\sigma$ & Zero with probability 0.75; otherwise log-uniform on $[0.01,0.1]$\\
\bottomrule\end{tabular}
\caption{Per-game dynamics parameters.}\label{tab:dynamics}
\end{table}

Starting states are uniform in a game-specific box: its centre $c$ has independent
coordinates uniform on $[-0.5,0.5]$, and its shared half-width $b$ is log-uniform
on $[0.05,0.5]$.

\subsection{Reward: a random strategy and a random outcome}
The \emph{strategy anchor} is a sampled $32\to64\to64\to4$ network with tanh
hidden layers and linear outputs. For a given state, it selects the action
with the highest output score. Following that
action gives zero anchor reward; any other action gives
$-1$. An independent $32\to64\to64\to1$ network $f$, with tanh after every
layer, scores the raw next state. The two components have equal weight:
\[
r_t=-\tfrac12\mathbf1\{a_t\ne\pi_{\rm anchor}(s_t)\}
       +\tfrac12f(s_{t+1}),\qquad f(s_{t+1})\in[-1,1].
\]
Total reward lies in $[-1,0.5]$. The outcome term can compensate for anchor
disagreement through the states subsequently reached, so the anchor is not
guaranteed to maximize total return over subsequent transitions. Neither reward
network is trained.

We hold each dynamics/outcome draw fixed while trying up to 32 independent
anchors. A strategy is rejected if any action exceeds 75\% of choices on 256
reset states.
The first passing anchor is retained;
if none passes, the candidate is rejected.

This deliberately designed family aims to provide a useful signal within our
budget. Rewards reading several past states and continuous actions remain
untested: we could not afford meta-training ablations of all alternatives.

\section{From candidates to a training bank}
We want games in which learning can outperform simple action patterns.
For each candidate, we train Double DQN \cite{double-dqn}
twice, with independent seeds, for 65,536 interactions per learner, then
evaluate its final greedy policy. The learners use the same network,
optimizer and exploration schedule as the DQN baseline described below,
but use the Double-DQN target.
The references are the best of four constant actions and the best repeating
action schedule, selected on 64 paired calibration episodes.

We compare each learner with each reference on 64 episodes and keep a game
only if each average advantage is positive in 95\% or more of 2,000 redraws
with replacement of 64 episodes among them.

To avoid relying on a single ranking, we selected sixteen games from the
first 48 passes:
\begin{enumerate}
\item \textbf{Eight with the largest advantage over simple policies.}
For each game, take the smallest observed mean gain across the two seeds and two references,
then choose the eight highest values.
\item \textbf{Four that passed the strategy-anchor comparison.}
Among the remaining games, identify those where both Double-DQN seeds also beat
the anchor in discounted mixed-reward return, using the same 64 paired
evaluation episodes and resampling test.
Five qualified, and we sampled four uniformly.
\item \textbf{Four random remaining passes.}
Sample four uniformly without replacement from the remaining passes,
broadening the mix without claiming uniform coverage of game space.
\end{enumerate}
All sixteen games passed a short Double-DQN screen, so the bank may favor the
related DQN baseline. On a game where following
the anchor already beats the constant and repeating policies, learning which
action pays immediately may suffice to pass. Only the four beyond-anchor slots
require more than matching the anchor, and even beating it does not establish
that exploration or long-horizon credit assignment was necessary. We have not
compared selectors through separate teacher-training runs.

\section{Learning and evaluating the rule}
We follow DiscoRL's discovery procedure \cite{discorl}, with smaller agents and
a smaller training budget. Thirty-two agents train in parallel;
each is replaced after a sampled lifetime of 100k, 250k, 500k or 1M
agent--environment steps (sampling weights $8:4:2:1$).
DiscoRL used 20M, 50M, 100M or 200M steps. We chose shorter lifetimes primarily
for budget reasons. We expected agents learning from small state vectors to need
fewer steps than agents learning from images, but did not establish convergence.

For generated games, the agent's shared network is
$33\to64$ (ReLU) $\to64$ (linear). Its outputs include four action
scores, converted to probabilities by softmax, and a 600-dimensional prediction
vector trained using the teacher's targets.

We meta-trained GREMLIN for 224.8M agent--environment steps: 181.8M to train
agents and 43.0M in separate rollouts to assess agent performance for teacher
updates. We clipped
each agent's meta-gradient to a maximum norm of 1 and averaged the resulting
per-agent Adam updates. Discovery
used about 616 GPU-hours on eight shared A5000s, including synchronization and
checkpoint transfers. Disco103 used 2,048 dedicated TPUv3 cores for 60 hours
\cite{discorl}; the hardware and workloads differ, so these are not comparable units of compute.
Even Disco57 reached its best reported rule at about 600M steps per Atari game,
or roughly 34.2B across 57 games, compared with our 224.8M agent steps \cite{discorl}.
DiscoRL trained convolutional agents directly from images, whereas our agents
received low-dimensional state vectors and were expected to learn with less experience.

For evaluation we freeze each rule and train fresh agents for 4M interactions,
with an early readout at 500k. We measure mean undiscounted episode return.
\emph{Initial} uses the teacher's initialization weights to train an agent; comparing
it with GREMLIN measures the change attributable to this discovery run.
A separate baseline does not learn at all: it draws each action uniformly,
independently at every decision. We estimate its mean over 1,024 independently
seeded complete episodes per game.

All three learned rules use Adam at 0.01. As in discovery, each batch contains
86 replayed and 10 fresh trajectories of 28 transitions, from a buffer of
up to 1,024 trajectories. DQN uses a 64/64 ReLU Q-network,
Adam at 0.0003, discount 0.99 and Huber loss, with one update per 16 steps.
It uses a 32,768-transition replay buffer, batches of 128 after 4,096 steps
of warm-up, and copies the target network every 2,048 steps. Its exploration
probability decays linearly from 1 to 0.05 over the first 32,768 steps.

\section{Results}
\subsection{The generated tasks supplied usable experience}\label{sec:generated}
We test all sixteen training games with two independent agent seeds per method
and a budget of 4M steps per learner, with 32 endpoint assessment episodes each.
Learned-rule agents sample actions; DQN selects
its highest-valued action. We also assess the fixed anchor under the complete
mixed reward.

The final system improves mean final return on all sixteen games.
DQN remains ahead on every training game: discovery used the
generated experience, but did not reach conventional-algorithm performance.
In this two-seed comparison, GREMLIN's mean final return exceeds Disco103's on nine of sixteen games.

\begin{figure}[ht]\centering
\includegraphics[width=\linewidth]{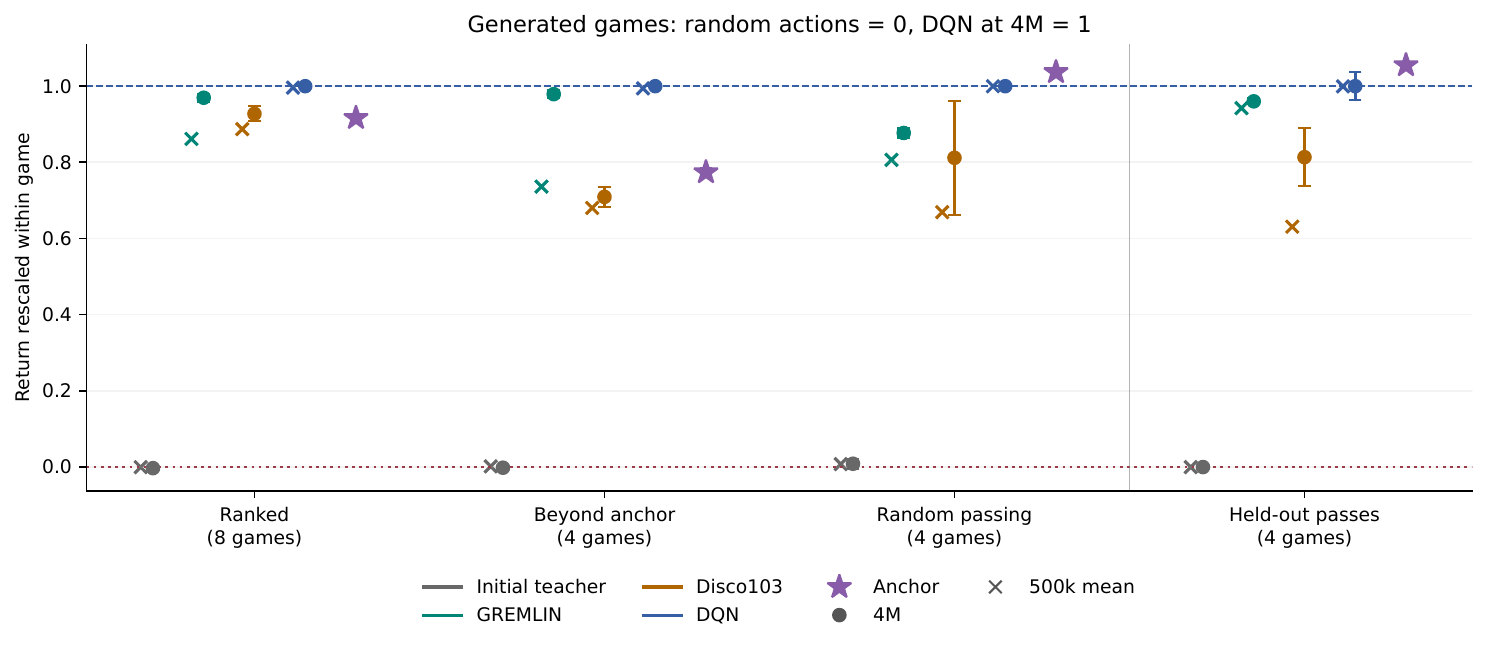}
\caption{The rightmost group was not used for discovery.
Circles and crosses average over games and agent seeds. The lower end
of each bar averages the lower of the two scores in each game; the upper end
averages the higher scores. These are not confidence intervals.}\label{fig:generated}
\end{figure}

In Figure~\ref{fig:generated}, zero means uniform-random actions and one means
DQN's mean return at 4M. Within each game we subtract the random-policy mean
and divide by the gap between DQN and random. This puts returns from games
with very different reward ranges on a common scale.

The initial teacher scores within 0.01 of random in all three group means,
as expected from a randomly initialized teacher. GREMLIN closes
97\% of the random-to-DQN gap on the ranked games, 98\% on the beyond-anchor
games and 88\% on the random passes. On the beyond-anchor games, DQN's advantage
over the anchor is expected from selection. GREMLIN also exceeds the anchor on
all four, but takes longer: its group score rises from 0.74 at 500k to 0.98
at 4M, whereas DQN is already
at 0.99 at 500k.

We repeated the comparison on four games drawn uniformly from the 32 screened
passes not used for discovery. GREMLIN improved over initialization on all
four, exceeded DQN's mean on one and averaged 96\% of the random-to-DQN gap.
GREMLIN scored comparably on the held-out and randomly selected training games;
this suggests generalization within the screened family, but says little about
how much of game space the generator covers. It may cover too narrow a range
of learning problems. The remaining gap to DQN on most
held-out games and all six human-designed games could reflect limited task
diversity, too little meta-training, or both.

\subsection{Human-task gains are uneven and below DQN}
We test the same four methods on Acrobot-v1 and discrete LunarLander-v3 using
native state vectors, and on Atari v5 Breakout, Space Invaders, Qbert and
Seaquest using RAM observations.
Each method has five independently initialized agents per game, trained for
4M agent steps on every task. Final scores
average 100 evaluation episodes per agent. Learned rules use sampled actions
and DQN is greedy, as in the generated comparison.

DiscoRL's published Atari evaluation \cite{discorl} uses a large CNN and four
image frames and 200M reported environment steps per learner.
Our Atari budget is 4M agent decisions
(nominally 16M emulator frames); step units are not assumed identical.
Our agents instead receive a single RAM vector (128 bytes scaled to $[0,1]$)
and choose from each game's own action set once every four emulator frames.
On each frame the emulator has a 25\% chance of repeating the previous action.
All six human tasks use the same dense agent architecture, with task-specific
input and action dimensions and no added clock.

\textbf{The improvement is not limited to generated training games.}
GREMLIN's mean final return exceeds both uniform-random play and Initial on
all six human tasks. On LunarLander and Seaquest, all five trained-rule agents
exceed all five
initial-rule agents. Space Invaders illustrates how unreliable this improvement
can be: three GREMLIN seeds score 271--284, near DQN's mean of 297,
while two score only 19 and 34, below the initial rule.

\textbf{The evaluation budget changes what learning is visible.}
GREMLIN is almost always unsuccessful on Acrobot at 500k steps (mean return
$-499$), but four of five agents learn to reach its goal by 4M (mean return
$-227$). On LunarLander its mean rises from $-53$ to $47$.
Disco103's Qbert mean also rises from 327 to 874. DQN appears close to a
plateau by 500k on Acrobot ($-88$ to $-82$) and Space Invaders (291 to 297),
but not across the panel: on LunarLander its mean rises from 76 to 188.

\begin{figure}[!ht]\centering
\includegraphics[width=\linewidth]{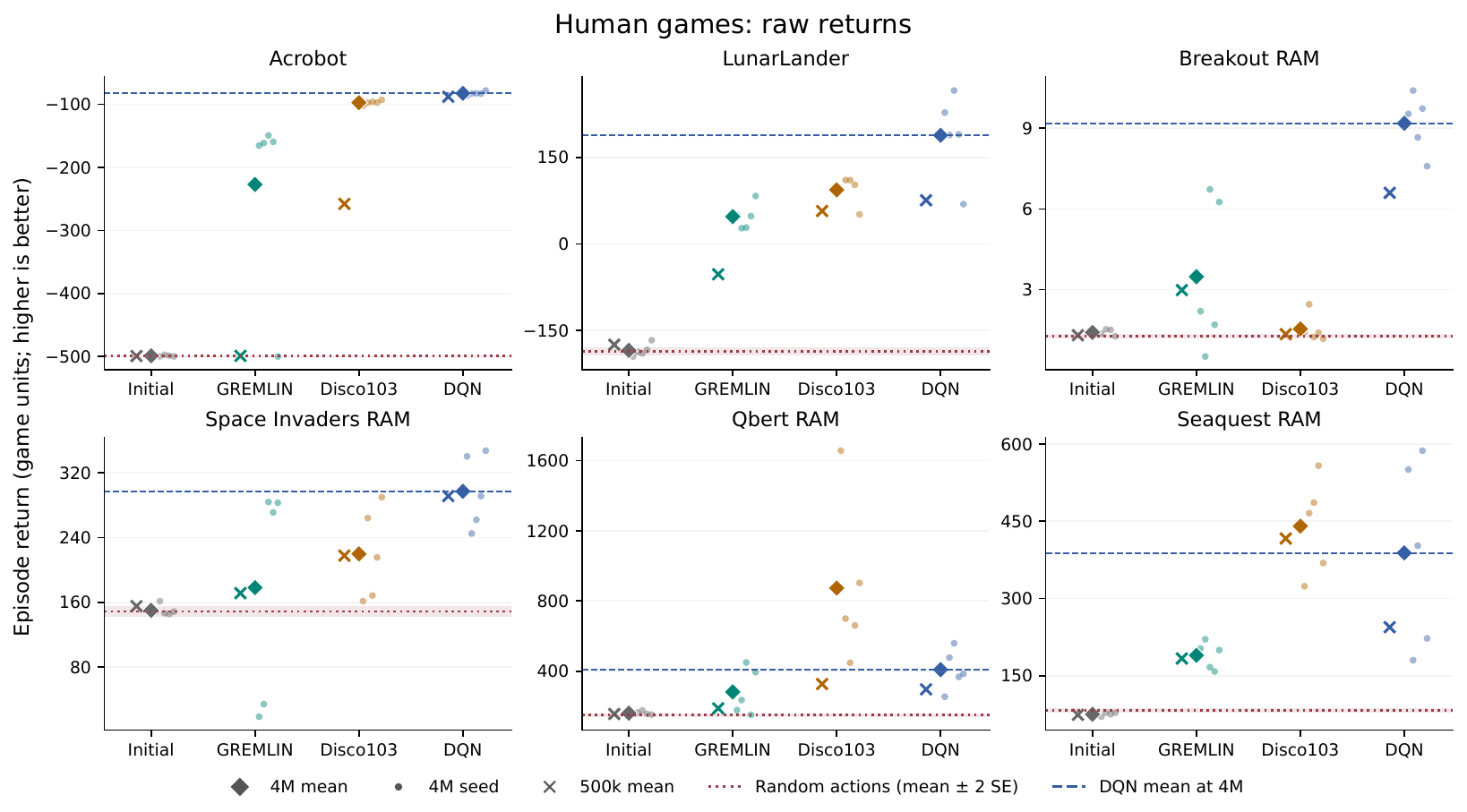}
\caption{Raw episode returns; each game has its own axis. Small points show
individual agents at 4M. The red band is $\pm2$ standard errors of the
uniform-random mean across 1,024 episodes.}\label{fig:human}
\end{figure}

GREMLIN remains below DQN's mean final return on all six human tasks and below
Disco103 on five; its mean exceeds Disco103's on Breakout. Disco103 itself exceeds
DQN on Qbert and Seaquest, so neither reference uniformly dominates the other.
These comparisons show gains outside the discovery bank, not an advantage
over established algorithms.

\begin{table}[!ht]\centering\small
\begin{tabular}{lrrrr}\toprule
Task & Initial & GREMLIN & Disco103 & DQN\\\midrule
Acrobot & \shortstack{$-498.72\pm1.02$\\$(-497.22)$} & \shortstack{$-227.08\pm152.66$\\$(-149.31)$} & \shortstack{$-97.19\pm3.40$\\$(-92.86)$} & \shortstack{$\mathbf{-82.49}\pm\mathbf{2.92}$\\$(\mathbf{-78.04})$}\\[3pt]
LunarLander & \shortstack{$-184.60\pm10.75$\\$(-166.94)$} & \shortstack{$47.38\pm22.64$\\$(83.09)$} & \shortstack{$93.66\pm24.71$\\$(110.59)$} & \shortstack{$\mathbf{188.23}\pm\mathbf{73.89}$\\$(\mathbf{265.91})$}\\[3pt]
Breakout RAM & \shortstack{$1.42\pm0.11$\\$(1.53)$} & \shortstack{$3.48\pm2.82$\\$(6.73)$} & \shortstack{$1.55\pm0.52$\\$(2.46)$} & \shortstack{$\mathbf{9.18}\pm\mathbf{1.09}$\\$(\mathbf{10.40})$}\\[3pt]
Space Invaders RAM & \shortstack{$149.96\pm6.46$\\$(161.30)$} & \shortstack{$178.08\pm138.73$\\$(283.90)$} & \shortstack{$219.77\pm56.89$\\$(289.80)$} & \shortstack{$\mathbf{297.13}\pm\mathbf{45.74}$\\$(\mathbf{347.30})$}\\[3pt]
Qbert RAM & \shortstack{$160.75\pm10.25$\\$(177.00)$} & \shortstack{$281.85\pm133.05$\\$(449.75)$} & \shortstack{$\mathbf{873.60}\pm\mathbf{466.52}$\\$(\mathbf{1656.50})$} & \shortstack{$408.95\pm115.57$\\$(559.25)$}\\[3pt]
Seaquest RAM & \shortstack{$74.88\pm3.37$\\$(78.20)$} & \shortstack{$189.48\pm26.32$\\$(220.60)$} & \shortstack{$\mathbf{440.52}\pm\mathbf{93.93}$\\$(\mathbf{558.00})$} & \shortstack{$388.52\pm185.14$\\$(587.20)$}\\\bottomrule
\end{tabular}

\caption{Raw human-task returns at 4M: mean $\pm$ sample standard deviation
over five agents, with the best agent in parentheses.}\label{tab:human-raw}
\end{table}

\FloatBarrier
\section{Limitations and research directions}
\paragraph{Task coverage.}
Our games reward agreement with a state-dependent strategy at every step,
alongside a random state score, and reset after 128 steps. Human episode limits
are 500 steps for Acrobot, 1,000 for LunarLander and 27,000 action decisions for
Atari; episodes can end earlier. Unlike our generated games, these tasks can
end in response to an action and have no added clock. The human tasks also
have 3--18 actions rather than four, and some have larger or sparser rewards
than our generated games.
What matters for transfer is whether discovery exposes the teacher to the
learning problems these tasks pose. In Acrobot, for example, an agent must
build momentum before it can reach the goal and end the per-step penalty.
Our dense rewards and short episodes may offer few examples where long-horizon
reasoning is needed. The late gain on Acrobot shows the rule can learn this
behaviour, not if it is efficient at doing so.

\paragraph{Training and agent capacity.}
The teacher may learn slowly or unstably. Separately, applying a useful rule
with an unsuitable agent learning rate or insufficient capacity can limit
performance.
A rule might improve from random initialization towards
basic competence without surpassing conventional algorithms.

\paragraph{Designed distribution and replication.}
Without meta-training ablations, we cannot attribute success to the anchor,
selection or inertia. Richer environments could help or make discovery harder:
this proof of concept does not establish that scaling the current design works.
Compute-matched discoveries on unscreened neural games and simple procedural
tasks would help distinguish the value of this generator from rule learning
on inexpensive tasks more generally.
An important comparison is breadth versus repetition: revisit a fixed bank or
keep supplying new games under the same selection rule, matching discovery
steps.

\section{Conclusion}
Deliberately structured random neural games supplied experience from which a
more useful learning rule emerged in this run.
DQN still led on most games, and the improvement could saturate below it.
Further experiments with a larger meta-training budget could show whether
GREMLIN continues to improve and how far that improvement carries over to
human-designed games.

\clearpage
\section*{Use of generative AI}
Generative AI tools assisted with software implementation, running
experiments, and writing the manuscript.

\bibliographystyle{unsrtnat}
\bibliography{references}
\end{document}